\documentclass[11pt,letterpaper]{article}
\usepackage{cogsys}
\usepackage[T1]{fontenc}
\usepackage{times}
\usepackage[pdftex]{graphicx} 
\usepackage{minted}
\usepackage{multirow}
\usepackage{booktabs}
\usepackage{xcolor}
\usepackage[hidelinks]{hyperref}
\usepackage{csquotes}

\usepackage{caption}
\usepackage{subcaption}

\usepackage{natbib}
\cogsysheading{12}{2025}{}{8/2025}{10/2025}

\ShortHeadings{CORGI: Efficient Pattern Matching With Quadratic Guarantees}
              {D.\ Weitekamp}

\begin{document} 

\title{CORGI: Efficient Pattern Matching With Quadratic Guarantees}
 
\author{Daniel Weitekamp}{weitekamp@gatech.edu}
\address{School of Interactive Computing, Georgia Tech, 
         Atlanta, GA 30332 USA}

\vskip 0.2in
 
\begin{abstract}

Rule-based systems must solve complex matching problems within tight time constraints to be effective in real-time applications, such as planning and reactive control for AI agents, as well as low-latency relational database querying. Pattern-matching systems can encounter issues where exponential time and space are required to find all matches for rules with many underconstrained variables, or which produce combinatorial intermediate partial matches (but are otherwise well-constrained). Coding tricks for writing production rules and database queries can remediate many of these issues in practice when rules are handwritten. However, when online AI systems automatically generate rules from example-driven induction or code synthesis, they can easily produce rules with worst-case matching patterns that slow or halt program execution by exceeding available memory. In our own work with cognitive systems that learn from example, we've found that aggressive forms of anti-unification-based generalization can easily produce these circumstances. To make these systems practical without hand-engineering constraints or succumbing to unpredictable failure modes, we introduce a new matching algorithm called CORGI (Collection-Oriented Relational Graph Iteration). Unlike RETE-based approaches, CORGI offers quadratic time and space guarantees for finding single satisficing matches, and the ability to iteratively stream subsequent matches without committing entire conflict sets to memory. CORGI differs from RETE in that it does not have a traditional $\beta$-memory for collecting partial matches. Instead, CORGI takes a two-step approach: a graph of grounded relations is built/maintained in a forward pass, and an iterator generates matches as needed by working backward through the graph. This approach eliminates the high-latency delays and memory overflows that can result from populating full conflict sets. In a performance evaluation, we demonstrate that CORGI significantly outperforms RETE implementations from SOAR and OPS5 on a simple combinatorial matching task.  

\end{abstract}

\section{Introduction}

Production systems (forward chaining rule systems) \citep{newell1973production} are the component within many cognitive systems that execute procedural knowledge (production rules) to carry out behaviors in reactive control and planning environments. Within AI agents, production systems use matching algorithms to determine which rules will be applied to carry out internal mental inferences and actions in external environments. The execution of a production rule adds or removes objects from working memory or executes an action in the world that is later reflected in working memory as the agent's environment changes. Matching systems also have many practical applications for retrieval, inference, and condition checking in database systems. 

Reducing matching times is essential to the effective operation of many forms of knowledge-based AI. Rule-based agents that operate in real-time environments must be able to determine their next action before time-sensitive opportunities have passed. In continuous control scenarios \citep{langley2022motion} like robotics and games, AI agents may even need to decide upon next actions timestep-by-timestep, necessitating latencies of tens of milliseconds or less. In database systems, low-latency and low-memory-footprint matching algorithms are essential for effectively utilizing computational resources and reducing delays experienced by end-users. 

Database systems' matching algorithms are typically designed for individual queries over large databases and employ simple matching approaches that are effective over large databases but are often ineffective for applying complex sets of rules. By contrast, production systems, like those used in cognitive architectures \citep{laird2019soar, anderson1997act} and similar symbolic AI agents, employ matching systems that test sets of complex rules simultaneously and match against comparatively small working memories that are updated as rules are executed one at a time.

\begin{figure}
    \centering
    \includegraphics[width=0.85\linewidth]{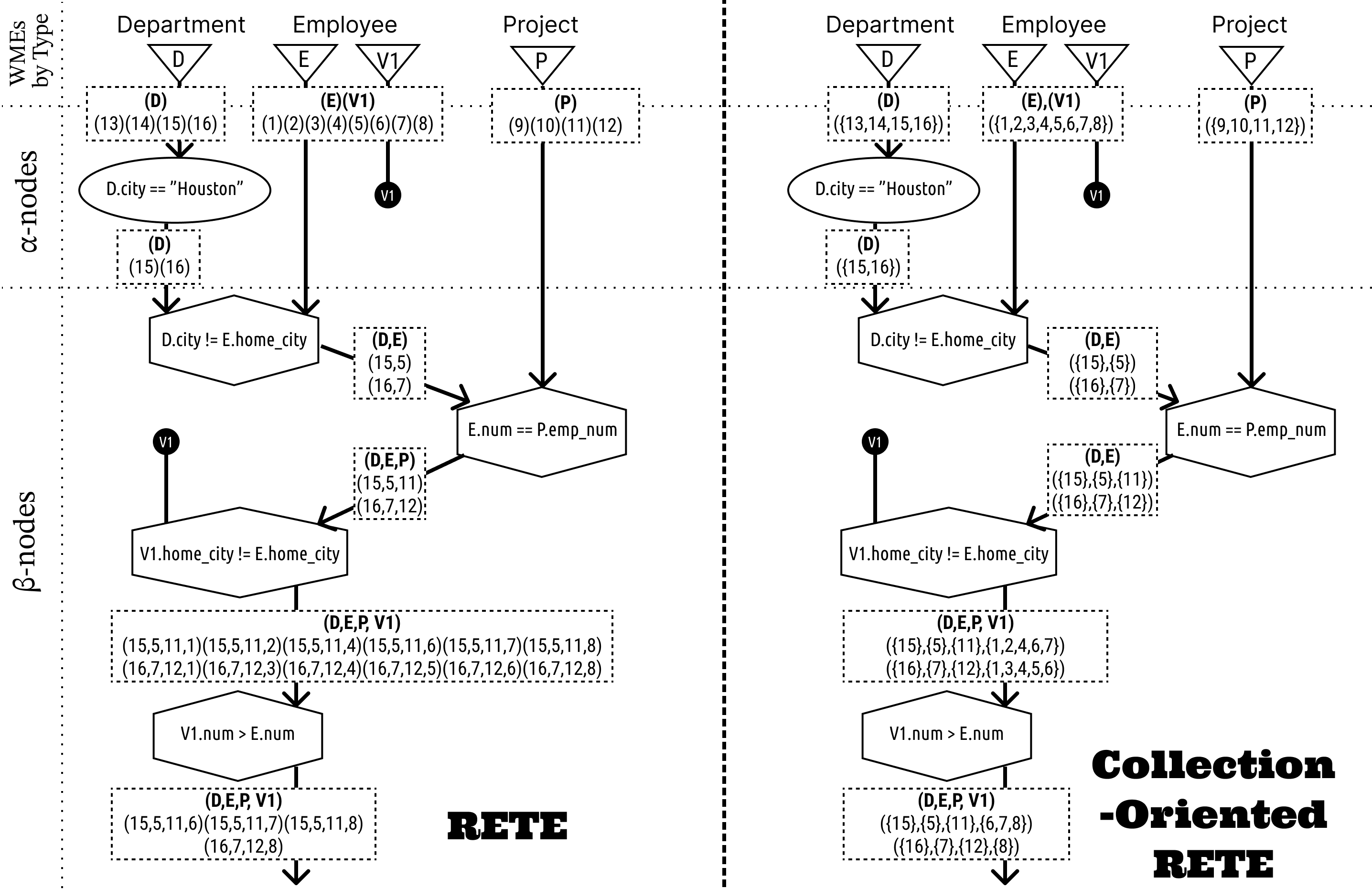}
    \caption{Graphs for RETE (left) and Collection-oriented RETE (right) applied to the data in Table \ref{tab:wm} on page 5 on the Valentine's task described on the same page. $\beta$-memories are shown in dashed boxes below each $\beta$-node in the lower third of the figure. Each integer is a WME identifier (see Table \ref{tab:wm}). RETE produces all combinations of matches, like (15,5,11,1), whereas in Collection-oriented RETE, combinations can be produced on demand by selecting WMEs from collections like \{11\} and \{1,2,4,6,7\}.
    }
    \label{fig:rete_vs_col}
\end{figure}

The vast majority of production systems employ variants of the RETE algorithm (see Figure \ref{fig:rete_vs_col}), which memoize intermediate match results with $\alpha$- and $\beta$-memories to maintain persistent conflict sets of candidate rule applications (i.e., bindings for rule arguments that satisfy their preconditions) \citep{forgy1989RETE}. Memoizing match results is an optimization tailored to the incremental nature of working memory changes in production systems. Between rule executions, many of the literals (i.e., predicates or their negations) within a ruleset's matching patterns will continue to have the same match candidates between match-execution cycles. So RETE updates the match results between cycles instead of recomputing them to speed up matching.

In RETE-based production systems, the match-phase can consume 90 percent or more \citep{forgy1981ops5} of the typical runtime, and many efforts have been made to optimize this performance-critical element of production systems. Early implementations of RETE  \citep{forgy1989RETE} relied on practices which modern software engineers generally consider anti-patterns in high-performance computing, such as heavy data copying and linear search over linked lists, leaving considerable room for improvement. Techniques such as hash map-based indexing \citep{nayak1988comparison} and multi-threading \citep{gupta1988parallel} have enhanced performance in numerous subsequent production system implementations. Some derivatives of RETE, like Treat \citep{miranker1987treat}, have explored alternative approaches to match cycle evaluations. However, almost all variations of RETE do not scale well to large datasets \citep{miranker1990performance} and suffer from the same exponential worst-case performance: $O(N^K)$ in memory and time, where $N$ is the working memory size and $K$ is the number of literals in a rule's preconditions. This combinatorial blowup is a risk in all derivatives of RETE that maintain sets of partial matches in $\beta$-memory. Partial matches are sets of working memory elements (WMEs; also sometimes called facts) that satisfy a subset of the literals in a rule's preconditions. Slowdowns and system crashes can occur when $\beta$-memories grow exponentially with different combinations of WMEs.     

Rules in production systems can sometimes be rewritten to mitigate these combinatorial blowups. For instance, by manually reordering the literals in rules' preconditions, one can sometimes influence how RETE graphs order join nodes that enforce relations between multiple objects in working memory. Some approaches have sought to address this issue automatically using heuristics to select between alternative join structures \citep{ishida2002optimization}, or unlink empty $\alpha$- or $\beta$-memories in multi-rule RETE graphs to avoid unnecessary checks \citep{doorenbos1993matching}. Other approaches compromise the consistency of the conflict set by focusing solely on maintaining match sets that maximize recency heuristics \citep{kang2004shortening}. Such shortcuts are appropriate for systems that use heuristics for planning, such as LEX \citep{mitchell1983learning}, but are inappropriate for systems that reason over all alternative matches. None of these approaches fully addresses the worst-case matching problems where the final conflict set (not just intermediate partial matches) is far too large to maintain in memory. For instance, a simple toy-problem scenario is a single rule that binds any $K$ different objects among the $N$ objects in working memory, resulting in a conflict set of size $N^K$. For instance, if N=100, K=5, and we are using an 8-byte addressing system, then RETE will have a memory overhead of at least $(5*8 \mathrm{\:bytes})*100^5= 400 \mathrm{\:GB}$ in this small worst-case scenario.  

Collection-oriented RETE \citep{acharya1993collection}  handles these situations more gracefully, although still without tight worst-case guarantees (see Figure \ref{fig:rete_vs_col}). Instead of holding all grounded partial matches at each join node, collection-oriented RETE's $\beta$-memory entries consist of collections of object identifiers $(e.g., \{1,2\}, \{3,6,7,8\}, \{9,17,24\})$, where each integer is a WME identifier. The collective terminal $\beta$-memories comprising the conflict set can be incrementally grounded into match sets, like $(1,3,9)$, on demand by simply iterating over different combinations within the Cartesian product of the collections; for instance, $(2,6,24)$ is another possible combination. However, collection-oriented RETE can only alleviate combinatorial explosions when the collections within a match entry are fully unconstrained. If a matching rule were to select $K$ \textbf{unique} objects among $N$ objects by imposing uniqueness with conditions such as \texttt{AND(A.id != B.id, B.id != C.id, A.id != C.id))}, then collection-oriented RETE's $\beta$-memory entries will "fragment" and suffer from the same worst-case combinatorial memory requirements as regular RETE. In practice, these worst-case scenarios can be alleviated by clever knowledge engineering. For instance, in the above example, one could divide a rule that selects K=5 unique objects into five rules fired in sequence that select the five unique objects one at a time, committing each selection to working memory.   

For veteran knowledge engineers, clever tricks can suffice to alleviate challenging matching situations. Yet tricky workarounds can compromise the expressivity and interoperability of production rules. In many modern applications of knowledge-based AI \citep{lawley2024val,kirk2014interactive}, such as interactive task learning \citep{itl}, where rules are learned or synthesized instead of hand-authored, there are no guarantees that generated rules will avoid matching situations that overwhelm RETE-based approaches. In our work on interactively teachable AI agents \citep{weitekamp2025decomposed,weitekamp2024ai2t,weitekamp2021toward}, we have found that agents very regularly produce rules with challenging matching patterns as part of their normal learning processes, and when they make reasonable inductive mistakes like trying to generalize two seemingly related (but in fact unrelated) rules into one through anti-unification, they can often produce worst-case matching patterns that are highly unconstrained, and would quickly crash conventional RETE-based approaches, in addition to collection-oriented RETE approaches.

\section{CORGI: Collection-Oriented Relational Graph Iteration}

In this paper, we present a new matching algorithm called CORGI (Collection-Oriented Relational Graph Iteration), which can reliably generate matches even in worst-case combinatorial scenarios and perform match cycles with tight quadratic guarantees in computation time and memory requirements. The essential difference between CORGI and RETE-based matching is that CORGI never commits partial matches or full conflict sets to memory. Instead, it maintains a relation graph that narrows down match candidates. The relation graph can be iterated over to generate match candidates as needed. Maintaining the relation graph and selecting a next match is $O(KN^2)$ where K is the number of objects in the matching pattern and N is the size of working memory\footnote{Linearity in K assumes a constant bound on the number of literals per object in each pattern, otherwise it is linear in L, the number of literals.}. However, of course, if a pattern has $O(N^K)$ matches, and for some reason all of them must be enumerated (which is usually unnecessary), then CORGI will have the same asymptotic performance as RETE.


We developed CORGI as part of a larger toolset called CRE (Cognitive Rule Engine)\footnote{ https://github.com/DannyWeitekamp/Cognitive-Rule-Engine/}, a high-performance Python extension that provides a toolset of algorithms for planning, matching, structure-mapping \citep{falkenhainer1989structure}, and generalization that are commonly used in cognitive systems. While many cognitive systems are derived from LISP, or implement LISP-like Domain-Specific Languages (DSLs), we have found these historical conventions to be difficult to maintain as part of a forward-looking research agenda. The go-to languages of modern software engineers (i.e., Python, C/C++/C\#, Java, and JavaScript) share syntactic conventions that are decidedly different from what historical cognitive systems employ. Our current CORGI implementation is written in Python and compiled with a just-in-time compiler called Numba \citep{lam2015numba}, enabling high performance within the convenience of the Python ecosystem.

\subsection{Working Memory and Matching Pattern Representation}

In general, the CRE toolset emphasizes first-class usage, where objects such as WMEs, conditions, and rules are written and manipulated directly in Python, rather than being loaded from a DSL and executed in a fixed manner (like executing a planner from STRIPS or a production system from OPS5). We've found that first-class interoperability with Python has been highly beneficial for the rapid development of AI systems, such as our interactive task learning agents \citep{weitekamp2020CHI, weitekamp2024ai2t}, which employ various symbolic learning and performance systems in combination. 

CORGI's working memory can be initialized as typed objects that have been predefined to have particular named attributes:

\begin{table}[h]
    \centering
    \begin{tabular}{rl}
       \texttt{1:} & \texttt{Employee(num=1, home\_city="Seattle", dept\_num=1)} \\
        \texttt{2:} & \texttt{Employee(num=2, home\_city="Orlando", dept\_num=1)} \\
        \texttt{3:} & \texttt{Employee(num=3, home\_city="LA", dept\_num=6)} \\
        \texttt{4:} & \texttt{Employee(num=4, home\_city="New York",dept\_num=6)} \\
        \texttt{5:} & \texttt{Employee(num=5, home\_city="LA", dept\_num=7)} \\
        \texttt{6:} & \texttt{Employee(num=6, home\_city="Houston", dept\_num=7)} \\
        \texttt{7:} & \texttt{Employee(num=7, home\_city="Orlando", dept\_num=8)} \\
        \texttt{8:} & \texttt{Employee(num=8,  home\_city="Houston", dept\_num=8)} \\
        \texttt{9:} & \texttt{Project(proj\_num=10780, emp\_num=8)} \\
        \texttt{10:} & \texttt{Project(proj\_num=10781, emp\_num=6)} \\
        \texttt{11:} & \texttt{Project(proj\_num=10781, emp\_num=5)} \\
        \texttt{12:} & \texttt{Project(proj\_num=10782, emp\_num=7)} \\
        \texttt{13:} & \texttt{Department(city="LA", num=1)} \\
        \texttt{14:} & \texttt{Department(city="New York", num=6)} \\
        \texttt{15:} & \texttt{Department(city="Houston", num=7)} \\
        \texttt{16:} & \texttt{Department(city="Houston", num=8)} \\
    \end{tabular}
    \caption{Example of typed objects in working memory.}
    \label{tab:wm}
\end{table}

Consider the following matching scenario over this data:
\begin{displayquote}

SuperDooper Corporation employs a large number of remote employees. This February, the Houston office has tasked all remote employees currently assigned a project with a fun cross-office engagement task: mail a Valentine's letter with candy to N junior employees (i.e., any employee hired after them, as indicated by a higher employee number than theirs) who are living in a different city. 

\end{displayquote}

\newpage

Here is one way of expressing this matching problem in CORGI:

\begin{minted}[fontsize=\footnotesize]{python}
D, E, P = Var(Department, 'D'), Var(Employee,"E"), Var(Project,"P")
conds = AND(D.city == "Houston", E.home_city != D.city, E.num == P.emp_num)
V1 = Var(Employee,"V1")
conds &= AND(V1.home_city != E.home_city, V1.num > E.num)
\end{minted}

The conditions \texttt{conds} are constructed with four variables each with predefined types (\texttt{Department}, \texttt{Employee}, or \texttt{Project}) that each bind to objects in working memory: the employee sending the valentine \texttt{E} (who's \texttt{home\_city} is not their department's city), their department \texttt{D} (which must be "Houston"), their project \texttt{P} (with \texttt{emp\_num} the same as their \texttt{num}), and the employee they are sending the valentine to \texttt{V1} (who is not in the same city and has a higher employee \texttt{num}). As first-class objects, the conditions in this example are constructed over several lines of code, but are equivalent to the following standard form, which can be visualized by printing the $\texttt{conds}$ object: 

\begin{minted}[fontsize=\footnotesize]{python}
AND(D:=Var(Department, "D"), D.city == "Houston",
    E:=Var(Employee,"E"), E.home_city != D.city, 
    P:=Var(Project,"P"), E.num == P.emp_num,
    V1:=Var(Employee,"V1"), V1.home_city != E.home_city, V1.num > E.num
)
\end{minted}

A notable feature of these expressions is that variables (i.e., \texttt{D,E,P,V1}) are bound to objects in working memory instead of to attribute values of those objects, and attribute values are dot-notation dereferences of objects (\texttt{D.city}). This is a feature of our implementation, and not a hard requirement, but we find that expressing literals in this way aids readability, is more consistent with typical Python expressions, and avoids the need to invent intermediate variable names. For instance, an equivalent expression of conditions in OPS5 would be:

\begin{minted}[fontsize=\footnotesize]{lisp}
(department ^num <d> ^city Houston ^city <dc>)
(employee ^num <e> ^dept\_num <d> ^home\_city {<c> <> <dc>})
(project  ^emp\_num {<=> <e>})
(employee ^home\_city {<b> <> <dc>} ^num {<v> <> <e>} )
\end{minted}

which is challenging to comprehend without careful study of the OPS5 manual, and contains many intermediate variables like \texttt{<dc>} which is bound to the \texttt{\^{}city} attribute of the department.

\subsection{Forward Pass: Updating the Relation Graph}

\begin{figure}
    \vspace{-1em}
    \centering
    \includegraphics[width=0.40\linewidth]{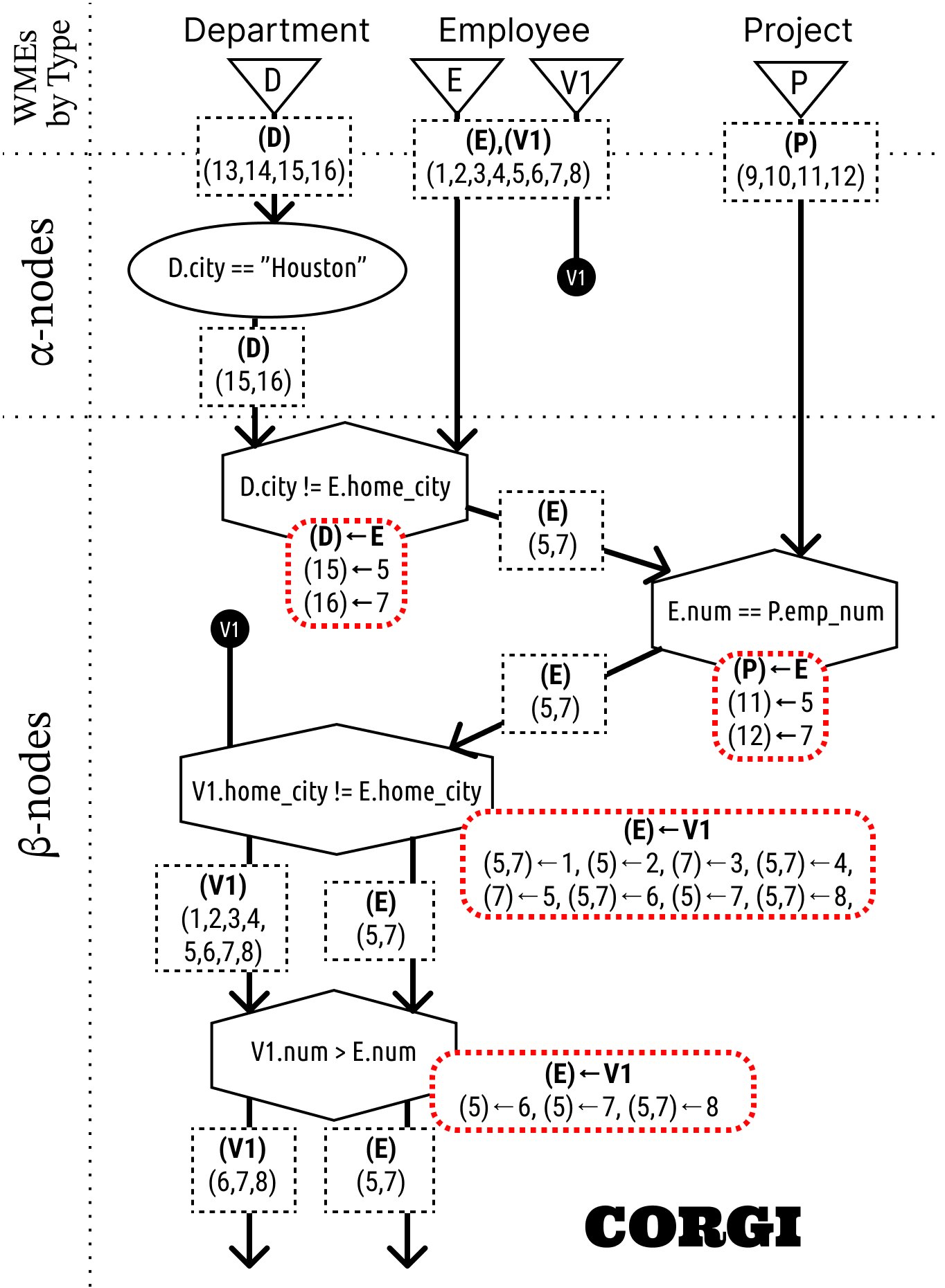}
    \caption{CORGI relation graph for Valentines problem using data from Table \ref{tab:wm}. Black dashed boxes show collections of WME identifiers in edges between nodes that are still candidates for bindings to a particular variable (e.g., \textbf{(E)}). Red dotted boxes show variable binding mappings (e.g. (11) $\leftarrow$ 5) for particular variables (e.g. \textbf{(P) $\leftarrow$ E)}).}
    \vspace{-1em}
    \label{fig:corgi}
\end{figure}

CORGI's relation graphs are similar to RETE graphs in that unary $\alpha$ predicates (e.g., \texttt{D.city == "Houston"}) and binary $\beta$ relations (e.g. \texttt{V1.num < E.size}) are handled by different kinds of nodes, with the $\alpha$ nodes filtering match candidates in advance of the $\beta$ nodes (join nodes). The principal difference between CORGI's relation graphs and RETE graphs is that $\beta$-nodes do not output sets of partial matches that can grow combinatorially deeper in the graph (see Figure \ref{fig:rete_vs_col}).

Instead, each $\beta$-node in CORGI is only responsible for maintaining a mapping $M(a_i)\rightarrow \{b_0,...,b_n\}$ between satisficing pairs of bindings of one variable in a relational literal to the other variable. For instance, for the literal \texttt{V1.num < E.num} it may map WMEs that bind to \texttt{E} to paired WMEs for \texttt{V1} (or vice versa) that satisfy the relation. These mappings (thick red dotted boxes in Figure \ref{fig:corgi}) have at most a quadratic update time and memory footprint. Tighter performance bounds are possible for the most common kinds of relations, like equality checking (e.g. \texttt{A.department == B.department}), where value indexing can eliminate the need to iterate over all input pairs (although pairwise checks are still necessary for derived relations like \texttt{V1.num > E.num}).


Each $\beta$-node passes information downstream only to descendants with a common variable. Each of these parent-child connections only maintains a collection of WMEs for the variable they have in common that contribute to at least one successful match of the parent literal. The only information maintained in the edges of the relation graph is an accounting of which objects in working memory have not yet been ruled out as part of at least one match, requiring further checking in downstream nodes. In other words, the links between all nodes, even $\beta$-nodes, are similar to RETE's $\alpha$-memories (black dashed boxes in Figure \ref{fig:corgi}), and permit no possibility of combinatorial explosion. The "CO" for Collection-Oriented in CORGI is a reference to the fact that, like collection-oriented RETE, these sets of WMEs in each edge are expressed as self-contained collections, without combinatorial expansion with other objects. 

Relative to RETE and its derivatives, CORGI's processes for updating mappings for each node and the collections maintained at the input-output boundaries along each edge are amenable to extremely efficient implementations. Node mappings don't necessarily need to be implemented using conventional indexing approaches, such as binary tree maps or hash maps, that incur lookup overhead. One approach is to use a bit matrix to implement a truth table for each relation evaluated over the inputs. This data structure is amenable to fixed memory modifications and fast position-based lookups. For instance, an equivalent truth table for the mapping in the lowest node of (Figure \ref{fig:corgi}) for its two input collections (V1) : (6,7,8) and (E) : (5,7), would be:

\begin{table}[h]

\centering

\begin{tabular}{cc|ccc}
  &     & \multicolumn{3}{c}{\textbf{(V1)}} \\
  &     & 6 & 7 & 8 \\
\hline

\multirow{2}{*}{\textbf{(E)}} 
  & 5   & 1 & 1 & 1 \\
  & 7   & 0 & 0 & 1 \\
\end{tabular}
\label{tab:bittable}
\caption{Truth table for bottom \textbf{(E) $\leftarrow$ V1} in Figure \ref{fig:corgi}}
\end{table}

These bit tables can grow and shrink as new objects are added and removed from working memory. Like conventional array-based lists, their allocated memory can be doubled along each dimension as they grow with working memory insertions to amortize the costs of re-allocating and copying. Data does not need to be moved when objects are retracted from working memory. Instead, the row or column associated with the removed object can be marked as empty and repurposed for the next insertion. This approach avoids unnecessary copies and keeps the table's underlying data contiguous, which is tremendously beneficial for maintaining data locality on the CPU's internal memory caches.


\subsection{Backward Pass: Relational Graph Iteration}

The "RGI" part of CORGI is Relational Graph Iteration. Relational Graph Iteration iteratively generates matches by working backwards through the relation graph. Iteration begins by selecting a variable binding from a collection in the output of a terminal node (i.e., any node with no downstream children). All valid matches will include an object in the outputs of a terminal node. To generate a full match starting from one of these selections, the iteration process follows all reachable paths backwards through the node mappings. 

Graph iteration requires that the variables in the source matching pattern are assigned a strict order. This ordering dictates how matching patterns are organized in the relation graph (i.e., relations with earlier variables are checked first), the direction mappings take (i.e., (earlier) $\leftarrow$ later), and the order that object bindings are selected during iteration (i.e., in reverse order). 

A natural ordering is the order in which variables are expressed in the matching pattern. However, sorting the variables in descending order of their relation degree, the number of other variables with which they share at least one relation (e.g., E in our example has degree 3), can provide a small performance benefit by culling down object candidates earlier in the relation graph.  

If there are two nodes with the same variable pair (like the bottom two nodes in Figure \ref{fig:corgi}), then the intersection of mapping results must be found to resolve the candidate bindings of the other variable. Mapping results are also intersected if two bound variables both have relations with an unbound variable. If the relation graph is separable (e.g., A depends on B, and C depends on D, but there is no relation across the pairs), then both connected graphs are iterated over in all combinations. 

Importantly, not all matches necessarily need to be enumerated during match iteration. CORGI simply returns an iterator object that maintains the iteration state and can be queried for the next match. Production systems typically only fire one rule per match-execution cycle and use heuristics or activation calculations to choose the best next candidate rule match to fire. Currently, CORGI's match iteration order depends on the variable order in matching patterns. The match iterators for later variables increment only after all candidate matches for earlier variables are exhausted. However, we conjecture that there are simple and efficient ways to impose heuristic or activation-based biases on relation graphs directly, influencing which matches are generated first, without the need to sort an enumeration of all matches by a selection criterion. This would make heuristic or competitive activation-based execution strategies practical even in worst-case matching scenarios. 


Relational Graph Iteration's iteration approach influences only the ordering of match candidates within single production rules, but imposes no functional constraints on the selection between individual productions. CORGI can efficiently report that it has a set of matches for a production rule before instantiating them all explicitly, meaning it doesn't prevent preference-based selection between productions like in SOAR, or activation-based competition between productions like in ACT-R. Additionally, in cases where it is desirable for match selection criteria to depend on particular objects in working memory---like, for instance, in ACT-R, individual chunks with specific retrieval costs or activation levels---there is the possibility of influencing match preferences within CORGI's relational graph iteration by sorting match candidates at each iteration level. In principle, there is also the possibility of including weighting criteria over pairs of objects in such calculations, like the connection strengths between connected chunks. However, tertiary or higher-order weights would be difficult to accommodate within Relational Graph Iteration without compromising its efficiency benefits. In any case, such selection criteria are rare. 

\section{Performance Evaluation}

In an expanded version of our Valentine example, we compare CORGI to RETE implemented in OPS5\footnote{Xiaofeng Yang's 2013 quicklisp port of 19-OCT-92 version of OPS5, (https://github.com/briangu/OPS5)} and the SOAR cognitive architecture\footnote{SoarSuite 9.6.4} \cite{laird2019soar}. We use a larger dataset than our original example, which consists of 26 employees, 12 projects, and 12 departments (a variation of the dataset used by \citep{kang2004shortening}), as shown in Appendix A. In this evaluation, we evaluate each matching approach on $N\in[1,5]$ unique Valentines instead of just one, and increase the number of objects in working memory by increments of 50, by duplicating the original dataset of 50 objects. Duplicate objects are given unique identifier attributes where appropriate. This task is designed to surface RETE's poor performance when rules' matching patterns target many objects with loose constraints between them. We include a \href{https://github.com/DannyWeitekamp/Cognitive-Rule-Engine/tree/main/examples/valentine}{\color{blue} GitHub link}\footnote{https://github.com/DannyWeitekamp/Cognitive-Rule-Engine/tree/main/examples/valentine} with source code for this evaluation. We conjecture that this task would also create high levels of fragmentation in collection-oriented RETE, leading to similar asymptotic performance as RETE. However, we were unable to locate a collection-oriented RETE implementation for comparison.

\begin{figure}
\centering
\begin{subfigure}{.5\textwidth}
  \centering
  \includegraphics[width=.98\linewidth]{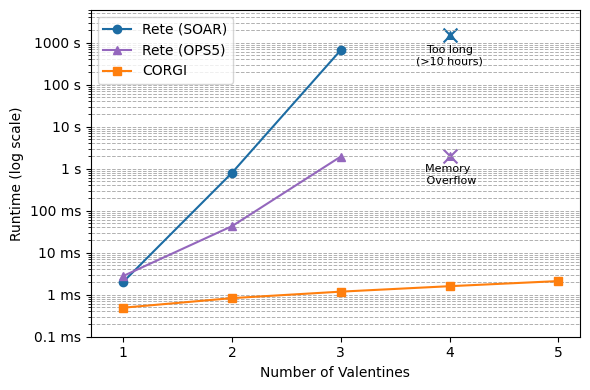}
\end{subfigure}%
\begin{subfigure}{.5\textwidth}
  \centering
  \includegraphics[width=.98\linewidth]{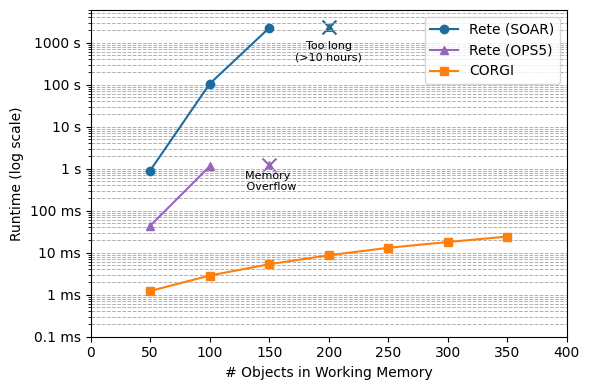}
\end{subfigure}
\caption{Log-scale runtimes for first match cycle of Valentine task (log scale) for SOAR and OPS5 implementations of RETE, and CORGI. (left) Varies the number Valentines in 1-5. (right) V=2 Valentines varying the number of objects in working memory. }
\label{fig:runtime}
\end{figure}

Figure \ref{fig:runtime} (left) shows the runtime\footnote{Run on Lenovo Yoga Pro 9 16IMH9 laptop with 32 GiB RAM, Intel® Core™ Ultra 9 185H × 22 processor.} of the first match cycle of the Valentines task for different numbers of Valentines. CORGI is considerably faster than RETE, and completes the matching task in 2 milliseconds or less for all numbers of Valentines. RETE's runtime suffers substantially as the number of Valentines increases. The OPS5 implementation takes 3.0 milliseconds at 1 Valentine, but a full 2.0 seconds at 3 Valentines, and fails completely with a memory overflow at 4 Valentines. SOAR's RETE implementation fares worse, with 3 ms, 800 ms, and 665 seconds for 1, 2, and 3 Valentines, respectively. Figure \ref{fig:runtime} (right) shows the runtimes for V=2 Valentines, with increasing working memory sizes. Similarly, the runtimes for OPS5 and SOAR increase dramatically as the number of objects in working memory increases. Meanwhile, CORGI's runtime remains quadratic in working memory size. Using AIC based-model selection, we verified that a quadratic model minimizes the AIC fit over a comparison of linear, quadratic, cubic, and exponential models.\footnote{The SOAR and OPS5 implementations fail too early along each variation, leaving too few points to identify their asymptotic behavior with curve fitting. However, by observation, they appear exponential.}    

At N=50 and V=4 Valentines, SOAR takes an impractically long time to complete the matching process. We waited for 10 hours before terminating it. SOAR is similarly unresponsive at V=2 Valentines and N=200 objects in working memory or greater. An inspection of SOAR's process memory at these points showed that it maintained a relatively low memory footprint of less than 1 gigabyte. SOAR's beta-memory implementation clearly stores candidate matches with a much lower memory footprint than conventional RETE implementations. Nonetheless, this does not appear to translate to a practical reduction in the runtime for producing match candidates for this simple combinatorial matching task. 

In both the OPS5 and SOAR implementations, RETE's performance deteriorates significantly as the combinatorics of the match problem increase. Meanwhile, CORGI's runtime increases marginally with the number of target objects in the matching pattern. The Valentine's task we use for this limited evaluation is considerably simpler than the matching patterns that we have observed to be induced by our interactive task learning systems. Albeit, those more complex patterns are often more well-constrained despite having dozens of variables, and so we typically see even shorter match times with CORGI in those cases. While we evaluate CORGI on just a single match cycle in this work, in future work, we plan to evaluate its performance as it is updated incrementally over multiple match cycles, as working memory is modified in full production systems. We expect to see similar performance benefits in these cases. 

\section{Conclusion}

Our motivation for developing CORGI was to create a solver for matching problems in cognitive systems that could operate under tight time constraints and under worst-case matching situations that permit enormous combinations of matched objects. In our prior work developing interactively teachable agents, we found that challenging matching situations arose quite often when executing the rules induced by our AI agents. We developed CORGI to provide a matching approach that would not break or introduce significant delays under these circumstances. As we've shown here, CORGI  is considerably more efficient than RETE-based approaches, generating matches in milliseconds that would crash RETE-based matching.

\vspace{-0.1in}

{\parindent -10pt\leftskip 10pt\noindent
\bibliographystyle{cogsysapa}
\bibliography{references}

}

\appendix

\section{Extended Data}

\texttt{\footnotesize
1: \ Employee(num=1, home\_city="Seattle", dept\_num=1) \\
2: \ Employee(num=10, home\_city="Orlando", dept\_num=1) \\
3: \ Employee(num=3, home\_city="Orlando", dept\_num=1) \\
4: \ Employee(num=17, home\_city="Orlando", dept\_num=2) \\
5: \ Employee(num=5, home\_city="Seattle", dept\_num=2) \\
6: \ Employee(num=19, home\_city="LA", dept\_num=2) \\
7: \ Employee(num=7, home\_city="Seattle", dept\_num=2) \\
8: \ Employee(num=8, home\_city="Dallas", dept\_num=3) \\
9: \ Employee(num=24, home\_city="LA", dept\_num=4) \\
10: Employee(num=2, home\_city="Dallas", dept\_num=4) \\
11: Employee(num=11, home\_city="Dallas", dept\_num=5) \\
12: Employee(num=26, home\_city="Dallas", dept\_num=5) \\
13: Employee(num=13, home\_city="LA", dept\_num=6) \\
14: Employee(num=23, home\_city="New York",dept\_num=6) \\
15: Employee(num=15, home\_city="LA", dept\_num=7) \\
16: Employee(num=16, home\_city="LA", dept\_num=7) \\
17: Employee(num=4, home\_city="Chicago", dept\_num=7) \\
18: Employee(num=18, home\_city="LA", dept\_num=8) \\
19: Employee(num=6, home\_city="LA", dept\_num=8) \\
20: Employee(num=20, home\_city="LA", dept\_num=8) \\
21: Employee(num=21, home\_city="LA", dept\_num=9) \\
22: Employee(num=22, home\_city="LA", dept\_num=9) \\
23: Employee(num=14, home\_city="LA", dept\_num=9) \\
24: Employee(num=9, home\_city="New York",dept\_num=9) \\
25: Employee(num=25, home\_city="New York",dept\_num=10) \\
26: Employee(num=12, home\_city="LA", dept\_num=10) \\
27: Project(proj\_num=10780, emp\_num=7) \\
28: Project(proj\_num=10781, emp\_num=8) \\
29: Project(proj\_num=10781, emp\_num=9) \\
30: Project(proj\_num=10781, emp\_num=1) \\
31: Project(proj\_num=10782, emp\_num=2) \\
32: Project(proj\_num=10782, emp\_num=3) \\
33: Project(proj\_num=10783, emp\_num=4) \\
34: Project(proj\_num=10784, emp\_num=5) \\
35: Project(proj\_num=10785, emp\_num=6) \\
36: Project(proj\_num=10785, emp\_num=10) \\
37: Project(proj\_num=10785, emp\_num=11) \\
38: Project(proj\_num=10786, emp\_num=12) \\
39: Department(city="LA", num=1) \\
40: Department(city="LA", num=2) \\
41: Department(city="LA", num=3) \\
42: Department(city="New York", num=4) \\
43: Department(city="New York", num=5) \\
44: Department(city="New York", num=6) \\
45: Department(city="Houston", num=7) \\
46: Department(city="Houston", num=8) \\
47: Department(city="Houston", num=9) \\
48: Department(city="Chicago", num=10) \\
49: Department(city="Chicago", num=11) \\
50: Department(city="Phoenix", num=12)
}

\end{document}